\documentclass[12pt, a4paper]{article}
\usepackage[font=footnotesize]{caption}
\usepackage[utf8]{inputenc}
\usepackage{graphicx}
\usepackage{authblk}
\usepackage{booktabs}
\usepackage{verbatim}
\usepackage{url}
\usepackage{caption}
\usepackage{subcaption}
\usepackage{indentfirst}
\usepackage{url}
\usepackage{amssymb}
\usepackage{amsmath}
\usepackage{array}
\usepackage{tcolorbox}
\title{Normalization in Proportional Feature Spaces}

\author{Alexandre Benatti$^1$ and Luciano da F. Costa$^2$}

\affil{
$^1$Institute of Mathematics and Statistics - DCC \\
University of S\~ao Paulo \\
Rua do Mat\~ao, 1010, \\ S\~ao Paulo, SP 05508-090 Brazil 
\\ \vspace{0.5cm}
$^2$S\~ao Carlos Institute of Physics - DFCM \\
University of S\~ao Paulo \\
Av.~Trabalhador S\~ao-Carlense, 400, \\ S\~ao Carlos, SP 13566-590 Brazil
}

\date{\emph{28th Aug., 2024}}

\begin{document}

\maketitle

\begin{abstract}
The subject of features normalization plays an important central role in data representation, characterization, visualization, analysis, comparison, classification, and modeling, as it can substantially influence and be influenced by all of these activities and respective aspects. The selection of an appropriate normalization method needs to take into account the type and characteristics of the involved features, the methods to be used subsequently for the just mentioned data processing, as well as the specific questions being considered. After briefly considering how normalization constitutes one of the many interrelated parts typically involved in data analysis and modeling, the present work addressed the important issue of feature normalization from the perspective of uniform and proportional (right skewed) features and comparison operations. More general right skewed features are also considered in an approximated manner. Several concepts, properties, and results are described and discussed, including the description of a duality relationship between uniform and proportional feature spaces and respective comparisons, specifying conditions for consistency between comparisons in each of the two domains. Two normalization possibilities based on non-centralized dispersion of features are also presented, and also described is a modified version of the Jaccard similarity index which incorporates intrinsically normalization. Preliminary experiments are presented in order to illustrate the developed concepts and methods.
\end{abstract}

\section{Introduction}\label{sec:introduction}

In data analysis (e.g.~\cite{cooley1971,dhar2013}), pattern recognition (e.g.~\cite{duda2000pattern,da2018shape,theodoridis2006pattern}) and scientific modeling (e.g.~\cite{gauch2003,meerschaert2013}), real-world entities are typically represented in terms of related numeric properties, measurements, characteristics, or \emph{features}, which often can be approached as random variables (e.g.~\cite{fisher1970,JohnsonWichern}), organized into respective \emph{feature vectors} inhabiting \emph{feature spaces}. This type of representation is necessary in order to allow objective quantitative methods to be employed for subsequent visualization, analysis, classification, and construction of models.

Interestingly, the representation of an entity in terms of some of its possible features constitutes itself a representational model of each entity, though typically not incorporating information about its dynamics and relationships with other entities. Indeed, establishing interrelationships between features and entities constitutes one of the main aspects in data and scientific modeling. An approach frequently employed to relate entities consists of \emph{comparing} their respective feature vectors by using some specific relationship operation (e.g.~\cite{costa2022simil}), such as distance or similarity indices. 

In case the comparisons are performed in terms of \emph{similarity} between pairs of feature vectors, it becomes possible to visualize the whole network of relationships between the original entities~\cite{da2022coincidence}. In this type of network representation of the interrelationships between the original entities, henceforth called \emph{similarity networks}, each original entity is represented by a respective node, while the similarities between the features of pairs of nodes are taken as respective weights.

In addition to similarity networks, it is also possible to consider complementary methods for data analysis and pattern recognition, including applying agglomerative clustering approaches~\cite{duda2000pattern,da2018shape,theodoridis2006pattern,benatti2024agglomerative}for estimating, in non-supervised manner, hierarchical interrelationships between the original entities as they are progressively merged into subgroups. 

Despite their potential for better representing, visualizing, characterizing, modeling and understanding real-world datasets and systems, approaches such as those mentioned above can be heavily influenced by the type of features adopted for characterization of the original entities to be studied, as well as by their respective \emph{normalizations} (e.g.~\cite{li2021,singh2015,singh2022,stolcke2008,liu2004,liu2007,xie2010}).

One particularly effective and general manner to approach the challenging topic of obtaining effective features in pattern recognition problems is in terms of \emph{feature transformations} (e.g.~\cite{sup_skewed}). Basically, given a feature $x$, which can be a preliminary feature or even a free variable, it is possible to use a function $f()$ so as to obtain a transformed feature $y = f(x)$. Interestingly, though derived from $x$, the new random variable $y$ is a feature on itself, which can be more or less effective for characterizing specific types of data elements while taking into account the context of each specific application.

The fact that both the random variables $x$ and $y$ mentioned above are features indicates that, given a feature $y$, it is important to know or try to infer, the initial feature $x$ from which it was derived through some transformation/normalization. Interestingly, each measurement of a physical property of a given entity can be conceptualized as a feature transformation from the real-world quantity $x$ into a respective measurement or feature $y$ with its specific physical unit. For instance, the measurement of the length of a leaf can yield a result in millimeters, centimeters, meters, or inches, among other possibilities. In addition to the adopted unit, the procedure of taking the measure can also imply intrinsic transformations (e.g.~logarithmic scale), noise, and error to the characterization of the initial property. All these aspects -- type of feature, adopted units, noise, sampling, and transformations -- can typically have a great impact on subsequent analysis of the data elements represented by feature vectors. Feature transformations are also frequently adopted as a means to modifying in some required manner a given feature, or for combining two or more previous features into a new feature.

Several of the methods typically applied to analyzing, classifying, and modeling data involve the \emph{comparison} between the properties of the supplied entities, represented in terms of a respective dataset. As discussed in~\cite{benatti2024agglomerative,sup_skewed}, there are at least two main types of comparisons: \emph{uniform} and \emph{proportional}. The former is characterized by translational invariance, while the latter is associated to scale invariance. Each of these two types of comparisons, which can be implemented in terms of a virtually infinite number of respective operators and similarity indices, have their intrinsic properties and relative advantages and limitations respectively not only to each specific problem and application but also regarding the types and characteristics of the involved features as well the questions and requirements being considered (e.g.~whether to give emphasis on comparisons between smaller or larger values). 

In the present work, an approach is developed which establishes a duality relationship between uniform comparisons taking place between uniform features and proportional comparisons between proportional features. This central result provides a principled approach to studying and trying to identify feature normalizations more intrinsically compatible with given type of features and comparisons. Though focus is placed on uniform and proportional features, it is also shown that several types of right skewed features can be treated in an approximated manner by several of the methods described in the present work. Left skewed features are not addressed in the present work.

It is argued that uniform features are intrinsically related to uniform comparisons as well as the normalization approach known as \emph{standardization}. Similarly, proportional features and comparisons are more directly related to normalizations considering the non-centralized dispersion of the negative and positive feature values. More specifically, two related approaches to the latter type of normalization are described which involve separated and joint consideration of the negative and positive values of proportional features. A modified Jaccard similarity index is also described which incorporates intrinsic feature normalization. However, it should be kept in mind that other normalization choices may be eventually possible as a means to implement specific requirements implied by specific problems (e.g.~to emphasize smaller or larger feature values).

A preliminary case example is also presented, illustrating the above aspects in which similarity networks are obtained in the case of a two-category dataset in both its uniform and proportional representations.

We start by discussing feature normalization from an integrated, ample perspective involving several other stages typically found in data analysis and modeling which can influence and be influenced by normalization. This is followed by a presentation the main concepts, properties, and methods underlying the normalization approaches described in the present work, including features, types of features, features transformations, as well as types of comparisons and their properties. Approaches to the normalization of the two main types of features addressed in this work, namely uniform and proportional, are then presented and discussed, followed by a complete case example involving similarity networks.

\section{An Integrated Perspective} \label{sec:integrated}

Though normalization is often considered from a more specific context aimed at equalizing or balancing some property of a given set of values, oftentimes it is actually part of a much larger perspective involving a whole research context underlying studies of which the normalization is a part. These additional aspects and choices include the original real-world (or abstract) system under study, how it can be approached and sampled, the properties (features) to be selected for the analysis, how they are going to be measured and then transformed and normalized, the methods to be employed for comparing and analyzing data elements, how data elements are to be interrelated in order to obtain respective models, as well as the main research questions and related requirements. 

\begin{figure}
  \centering
 \includegraphics[width=1 \textwidth]{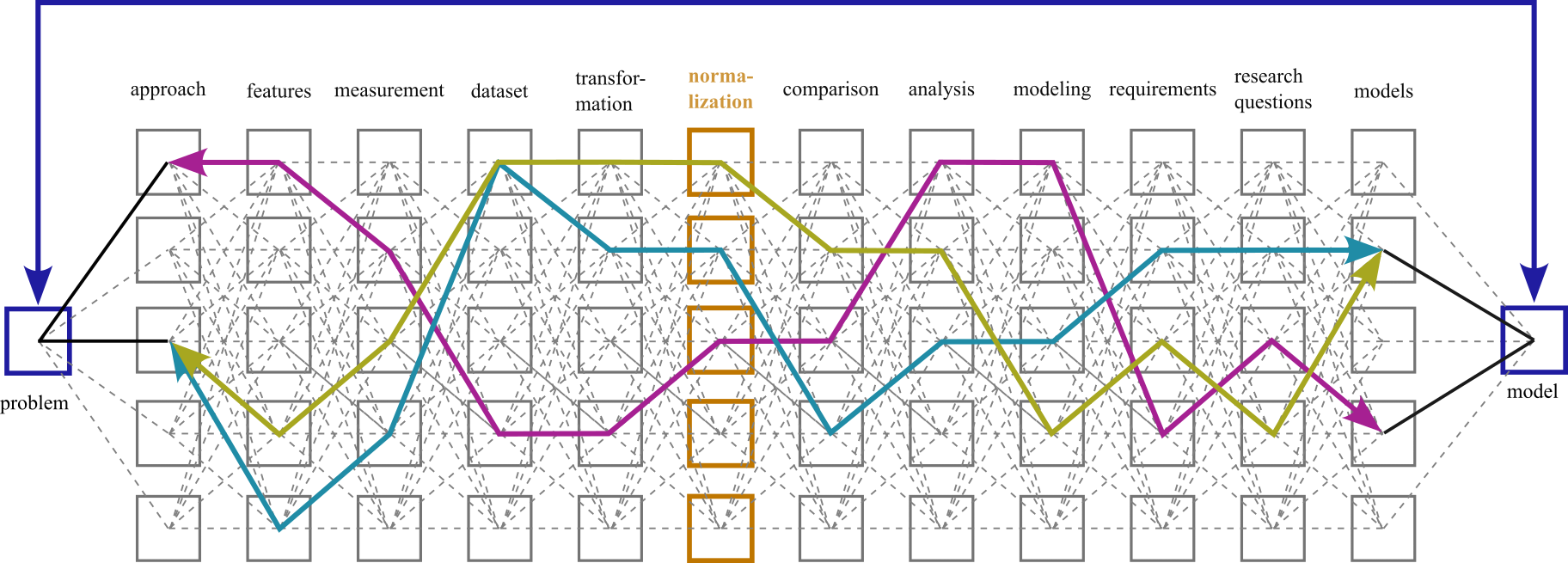}\\
 \caption{Diagram illustrating some of the main stages, shown as columns, typically involved in data analysis and modeling, defining the context in which normalization (emphasized near the middle of the diagram) is often approached. Also depicted are three possible approaches, represented as respective pathways. Each of these pathways can be obtained by making choices along each of the successive stages. The choice between the several possible pathways could consider, for instance, their respective effectiveness in obtaining a resulting modeling which is as stable and accurate as possible. However, each choice is influenced and influences all the other stages.}
 \label{fig:normalizing}
\end{figure}

Figure~\ref{fig:normalizing} presents a possible diagram illustrating the several possible stages and alternative approaches concerning some of the several aspects (shown as columns) typically involved in data analysis and modeling. Though all these aspects can interact one another (defining a completely interconnected graph of interrelations), specific approaches need to be delineated in terms of successive choices along several stages, represented as columns. Observe that, in general, other stages can be incorporated, and also that the columns be arranged in different orders. Observe that the structure in the figure is only a particular abstraction of a possible configuration of stages.

The objective, represented by the upper double-sided connection in black, is to relate the given problem represented by the block at the right-hand side of the diagram, with a suitable model shown as the block shown at the right-hand side. A possible solution can be understood as a \emph{pathway} (or `trajectory') passing over alternative possibilities at each successive column. Not all these choices are particularly effective or appropriate, and some could lead to overall inconsistencies and should therefore be avoided. Moreover, the choices at a given column typically relate to choices at other columns. For instance, the choice of type of comparison will directly influence and be influenced by the type of features and respective normalizations. 
Three possible solutions are illustrated as pathways in the diagram, each of which involving distinct choices among the several possible aspects. Observe that, though the yellow and blue pathways have common extremities, they follow distinct paths along the successive stages.

The choices along the possibilities in the normalization column are typically more related to the adjacent aspects, namely feature transformation and comparison, but these are then transitively related other successive columns, to the extent that all choices along all columns ultimately turn out to be all intrinsically interrelated. Therefore, the selection of a normalizing approach should ultimately take into account all the aspects in a data analysis or modeling problem. Though normalization constitutes only one among the several involved stages, it should be effectively addressed because it can influence and be influenced by virtually all the other stages.  A few of the possible interrelations between normalization and the other stages are briefly discussed as follows.

Going from left to right, starting from the problem of interest (e.g.~studying a given natural phenomena, or developing a framework for identifying specific data categories), we have that oftentimes there are several approaches to organizing, implementing, and sampling the experimental aspects, which can take place respectively to varying environments, duration, and initial conditions, among many other possibilities. This first stage already has implications on the normalization, because an experiment in varying conditions is likely to yield measurements with wider ranges of magnitude variation, or even involving changing statistical properties (e.g.~linear and exponential). Then, it is necessary to select which variables (or features) will be adopted for representing the aspects of interest specific to each experiment. Different variables are likely to have specific distributions of respective magnitudes, which will need to be considered while of their normalization. In the present work, we shall focus on two main types of features, which will be called \emph{uniform} and \emph{proportional}, but it should be kept in mind that there is a virtually infinite number of types of features with their specific properties.

Next, measurements have to be obtained respectively to each of the adopted variables, in order to yield respective datasets, and the own ways in which these measurements are made can strongly affect the properties of the obtained values. For instance, some measurements are implemented in logarithmic scale, and it is also possible that too large feature values can lead to saturation, all of which impacting on the way impact on the normalization procedure.  Typically, these measurements are subsequently organized into datasets, in which each data element is represented in terms of several respective features, which are often transformed in order to achieve some specific requirement, such as changing units, filtering noise or changing scales, which can again impact the normalization and any other stage. These transformations can also involve combinations of the original features into new features with possibly new properties. At this point, when the normalization stage proper is reached, the types and statistical properties of the originally chosen variables may have changed substantially.

The normalization stage is typically followed by comparisons between the available features. As discussed in the present work, there are several types of comparisons, so that it becomes interesting to chose an alternative which is not only compatible with the types of available features, but also with the way in which the comparisons are expected to work, while reflecting the research questions and requirements. For instance, it may be the case that the comparisons need to be more strict when comparing similar data elements, or that larger importance is to be assigned to certain ranges of feature magnitudes, all of which again influence the normalization procedure as well as other stages.

Similar needs are then likely to be imposed by the analysis and modeling stages, and these are too ample and diverse to be contemplated here. However, when moving from right to left, starting from the sought model, we also have that several established distinct hypothesis and \emph{research questions} will also need to be taken into account. For instance, emphasis can be placed on specific properties of the studied problem, such as robustness, long or short term predictions, choice between absolute or relative magnitudes, consideration of time and/or spacial scales, as well as proper recognition of specific structural or dynamical categories, among several other possibilities. Each of these choices will typically be accompanied by respective requisites, such as levels of accuracy and/or use of gold standards to evaluate the accuracy of the obtained results, to name but a few possibilities. All of this will imply additional constraints and aspects to be considered while developing and/or selecting an effective normalization procedure.

As could be expected, in general there is no predetermined manner to approach at once and definitively all the involved choices and aspects. Figure~\ref{fig:normalizing} illustrates three possible approaches represented in terms of respective pathways shown in light blue, yellow and magenta. Provided each of such possible solutions can be evaluated in terms of the quality of the respectively obtained results, it becomes possible to approach the issue of addressing data analysis and modeling by starting with some carefully defined preliminary pathway, to be progressively optimized in terms of the quality of the obtained results as well as the consideration of new and/or additional experiments and hypothesis. It is interesting to observe that a pathway found to allow good results respectively to a given data analysis and modeling problem will not necessarily be suitable for a distinct research problem.

All in all, the above discussion indicates that, typically, normalization is not an isolated activity to be approached in independent and absolute manner, though it is frequently (but not always) expected not to change the type (e.g.~uniform, proportional, etc.) of the supplied features. Instead, not only normalization, but actually all stages along a data analysis and modeling stages, are inherently interrelated and integrated into a much larger framework.

\section{Basic Concepts}

Several of the main concepts and methods employed in the present work, including features, feature spaces, types of features, as well as comparisons and some of their types, relationships and transformations are described and discussed in the following subsections. Also included is a description of multiset approaches to similarity indices, including a modified version of the Jaccard similarity index which incorporates inherent normalization. This section ends with a note discussing the issue of feature identification and transformation from practical perspectives.

\subsection{Features and Feature Spaces}

Given a set of original \emph{entities} or \emph{data elements} $k=1,2, \ldots, N$, they are assumed to be representable in terms of $M$ respective \emph{features} $x_{i,k}$, $i=1, 2, \ldots, M$, which can be organized into the following \emph{feature vector}:
\begin{align}
   \vec{x}_k = \left[ x_{1,k}, x_{2,k}, \ldots, x_{M,k} \right]
\end{align}

The following properties of each of the features $x_k$ can be considered:
\begin{align}
   & \emph{minimum value:} \ x_{min,k} = \min \left( x_{i,k}\right), i= 1, 2, \ldots, N \\
   & \emph{maximum value:} \ x_{max,k} = \max \left( x_{i,k} \right), i= 1, 2, \ldots, N \\
  & \emph{average value:} \ \mu_{x_k} = \frac{1}{N} \sum_{i=1}^{N} x_{i,k} \\
  & \emph{standard deviation:} \ \sigma_{x_k} = \sqrt{ \frac{1}{N-1} \sum_{i=1}^N \left(x_{i,k} - \mu_{x_k} \right)^2 }
\end{align}

Although feature vectors can be understood to belong to a respective M-dimensional \emph{feature space} $X$, in general situations it cannot be guaranteed that the axes defined by the original features are orthogonal, or even that inner product and metric can be consistently specified. Actually, as features often have distinct physical units, their linear combination, as is the case of inner products and distances, will frequently lead to inconsistent physical unities. This limitation can be partially addressed by \emph{normalizing} the original features so that they become dimensionless.

Even when dimensionless new features are obtained by normalizing the original counterparts, there is no guarantee that their representation as an orthogonal system of coordinates, as well as respective inner products and Euclidean distances, will actually be consistent from the point of view of the nature of the original measurements. For instance, as the properties of pressure, volume, and temperature at a given point of a gas have distinct physical units, there is not an immediate physical interpretation of their linear combinations (especially addition), orthogonality or independence, which may actually depend on specific problems of interest. At the same time, representing each of these features as respective orthogonal axes does not mean that the respectively implied features are independent random variables associated to system of coordinates underlain by a canonical basis.

For the above reasons, it becomes interesting to consider combinations of features in terms of basic operations such as union and intersection, which neither depend on linear combinations of the original features and nor require the axes to be orthogonal. Actually, all that is required is that each of the features correspond to a set (or, more precisely, multiset, e.g.~\cite{da2021multisets,da2024integrating}).

Approaching a feature $y$ in terms of the transformation of an invertible uniform or symmetric feature $x$, i.e.~$y = f(x)$, constitutes a particularly interesting approach since it allows the properties of $y$ to be more easily inferred from the properties of $f(x)$. This is adopted throughout this work.

Given a generic feature $x$ defined by its respective density probability (e.g.~\cite{JohnsonWichern,fisher1970}) $p_x(x)$, it can be shown that a new transformed feature $y= f(x)$ will have respective density $p_y(y)$ specified as:
\begin{align}
   p_y(y=f(x)) = \frac{1}{f'(x)} \, p_x(x),
\end{align}  \label{eq:transf}
where $f'(x)$ is the first derivative of $f(x)$.

As a consequence, feature transformations by monotonic functions $f(x)$ which have up concavity will be respectively characterized by right skewed densities, while down concavity transformations will lead to left skewed densities.

A multivariate \emph{separable function} is so that it can be expressed in terms of the product of univariate separated functions. Separable multivariate feature densities can be obtained by the product of univariate densities obtained by transformation of respective univariate densities. It is also possible to consider a multivariate form of Equation~\ref{eq:transf} to model more general multivariate densities.

Interestingly, the transformation of a random variable can be used to model measurements of physical quantities, being also directly related to expressions underlying respective models in terms of free variables. It is interesting to observe that linear measurements could be employed in order to achieve maximum fidelity while measuring a physical property. However, real-world measurement systems are almost invariably characterized by some level of non-linearity implied not only by sensors but also by amplifying and conditioning stages. In addition, non-linear (e.g.~logarithmic) scales have been adopted when dealing with physical quantities characterized by wide ranges of magnitude variations.

Though theoretically it is possible to consider an elementary feature in an absolute manner, by considering $x$ to be simply a free variable and not a preliminary feature, it turns out that in the case of real-world entities it is virtually impossible to approach a given feature without relating it to some other more fundamental counterpart. Interestingly, even the problem of identifying a definitive set of fundamental physical properties has not been completely defined, remaining an issue related to the interesting area of philosophy of science.

\subsection{Types of Features}

Henceforth, we limit our attention to numeric features represented numerically in terms of integer or real values.

A \emph{uniform feature} is so that its respective density is \emph{uniform}.

A \emph{proportional feature} has a density resulting from a transformation of the type $f(x) =c^x$ of a uniform feature $x$, where $c \in \mathbb{R}$, $c>1$. This type of transformation is henceforth called \emph{proportional}.

It is henceforth assumed that $\alpha \neq 0$ is a real constant. A \emph{proportional feature} $y$ should correspond to a result from a uniform (linear) transformation $y=f(x)$ of the free variable $x$ so that: 
\begin{align}\label{eq:unifeat}
    f( x + \alpha) - f(x) = f( \alpha ).
\end{align}

Observe that the variable $x$ has been taken as a free reference feature corresponding to a uniform density along a given interval $[a,b]$ along the axis of real values, i.e.$x \in \mathbb{R}$, with $p(x)=1/(b-a)$. In other words, $x$ can be said to be the reference uniform feature par excellence.

In other words, Equation~\ref{eq:unifeat} means that the difference between the uniform transformation of a feature $x$ translated by a constant $\alpha$ and the same transformation of $x$ is necessarily identical to the same transformation of $\alpha$.

Examples of transformations of $x$ leading to uniform features include:
\begin{align}
   &y = f(x) = 2 \, x \nonumber \\
   &y = f(x) = -3 \, x \nonumber \\
   &y = f(x) = \pi \, x \nonumber 
\end{align}

Observe that functions of the type $y = a\, x + b$, $a,b \in \mathbb{R}$, $a,b \neq 0$, do not yield uniform features because:
\begin{align}
f(x+\alpha) - f(x)= a \, (x + \alpha) + b - a\, x - b =
a \, \alpha \neq a\, \alpha + b = f(\alpha) \nonumber
\end{align}

A \emph{proportional feature} $y$ should correspond to the result from a proportional (non-linear) transformation $y=f(x)$ of the free variable $x$ which acts so that: 
\begin{align}\label{eq:propfeat}
    \frac{y_{x+\alpha}}{y_{x}} = \frac{f( x + \alpha )}{f( x ) } = f(\alpha) = y_{\alpha} 
\end{align}

The above characterization in terms of fractions is the main motivation for adoption of the adjective \emph{proportional} in both the terms \emph{proportional features} and \emph{proportional comparisons}.

We henceforth adopt the convention that $y_{x+\alpha} = f(x+\alpha)$, $y_{x} = f(x)$, etc.

Therefore, proportional features are analog to uniform features, but here it is the division of the transformed features, instead of the difference in Equation~\ref{eq:unifeat}, which is considered.

This property is immediately satisfied by the power function $f(x)=c^{ x}$, where $c \in \mathbb{R}$, $c > 1$, i.e.:
\begin{align}
    \frac{ c^{ x + \alpha}} {c^{ x} } = 
    \frac{ c^ x c^\alpha} {c^{ x} } = 
     c^{\alpha} 
\end{align}

In the case of proportional features, we have that:
\begin{align}
   &y_{x+\alpha} = c^{x+\alpha} = c^{\alpha} \, c^x = \beta \, c^x = \beta \, y_x \label{eq:beta}
\end{align}

Thus, we can write:
\begin{align}
   &y_{x_1+ \alpha} = c^{\alpha} \, y_{x_1} = \beta \, y_{x_1},\\
   &y_{x_2+ \alpha} = c^{\alpha} \, y_{x_2} = \beta \, y_{x_2}
\end{align}

which shows that \emph{translations in the uniform domain correspond to scalings in the proportional domain}. 

This result, which is specific to proportional features, indicates that under these circumstances the variation represented by the interval $[y_{x_1}, y_{x_2}]$ also scales linearly with $\beta$, i.e.:
\begin{align}
  y_x \in [y_{x_1},y_{x_2}] \Longrightarrow \beta \, y_{x} \in \left[ \beta\, y_{x_1}, \beta \, y_{x_2} \right] \label{eq:propintvl}
\end{align}

This property is characteristic of several real-world and abstract situations, more precisely those following power laws, which include the sizes of power outages, moon craters, and many other possibilities (e.g.~\cite{clauset2009}). This also seems to be the case with the size of animals or plants~e.g.~\cite{bonabeau1999}). As hinted by Equation~\ref{eq:propintvl}, proportional features can also be approximately related to features whose dispersion tends to scale linearly with the respective average.

For generality's sake, it is also interesting to consider \emph{right skewed features} which, though not corresponding to proportional features in the sense of the above described approach, share with them the property of being described by right skewed densities.

\subsection{Types of Comparisons}

A \emph{uniform comparison} is such that:
\begin{align}\label{eq:lin}
  \mathcal{R}_u \left(x + \alpha,y + \alpha \right) = 
    \mathcal{R}_u \left(x, y \right).
\end{align}

As an illustration of a uniform comparison between the values of two features $y_1$ and $y_2$, it is possible to simply make:
\begin{align}
    \Delta(x,y) = y-x.
\end{align}

In this case, the result of the comparison can be positive, negative, or zero, indicating the direction of the difference between the two values.

It is also possible to have the following uniform comparison:
\begin{align}
    \Delta_a(x,y) = |y-x|,
\end{align}

which can only take non-negative values. 

Let $y$ be a uniform feature obtained from a free variable $x$ in terms of the linear transformation $y=f(x)$, as well as its following four values $y_{x_1}$, $y_{x_2}$, $y_{x_1 + \alpha}$, and $y_{x_2+\alpha}$. The uniform comparisons $\mathcal{R}(y_1,y_2)$ and $\mathcal{R}_u(y_3,y_4)$ can be related as follows:
\begin{align}\label{eq:relunif}
   \mathcal{R}_u(y_{x_1+\alpha},y_{x_2+\alpha}) = 
   \mathcal{R}_u(y_{x_1}+y_{\alpha},y_{x_2}+y_{\alpha}) = 
   \mathcal{R}_u(y_{x_1},y_{x_2}). 
\end{align}

In other words, the results of the uniform comparisons between two values of a uniform feature and between these values translated by the same amount $\alpha$ are identical.

It is important to keep in mind that neither $\Delta$ or $\Delta^a$ are better or worse in absolute terms, as there will be situations in which each of them could be preferable.

Possibly, the \emph{Euclidean distance} corresponds to the most frequently employed uniform comparison between two M-dimensional feature vectors $\vec{x}$ and $\vec{y}$. The Euclidean distance can be calculated as:
\begin{align}
  E(\vec{x}, \vec{y}) = E(\vec{x}, \vec{y}) = | \vec{x} - \vec{y} |= \sqrt{ \sum_{k=1}^M \left(x_{k}-x_{k} \right)^2 }
\end{align}

The Euclidean distance is a uniform comparison invariant to translation, rotation, and inversion (i.e.~$\hat{x}_k = -x_k$) of any of the involved axes in the respective feature space. It should also be kept in mind that the Euclidean distance between two feature vectors require, for consistency's sake, that all elements of the vector have the same physical unit. Actually, the resulting 
Euclidean distance value will have the same unit as that of all the vector elements. Thus, it will only be dimensionless in case all features are dimensionless.

Being a uniform comparison, the Euclidean distance satisfies the following relationship regarding the translation of the two compared vectors by the same relative distance $\vec{\alpha}$:
\begin{align}
   E(\vec{x}+\vec{\alpha}, \vec{y+\vec{\alpha}}) =
   E(\vec{x}, \vec{y})
\end{align}

where $\vec{\alpha}$ is a vector of non-zero constant values $\alpha_k$.

It is also interesting to observe that a uniform comparison between two values $x$ and $x$ scales linearly with $\alpha \in \mathbb{R}$, $\alpha \neq 0$ when applied to the two arguments scaled by the same factor $\alpha$, i.e.:
\begin{align}
  \mathcal{R}_u \left( \alpha \, x, \alpha \, y \right) = 
    \alpha \, \mathcal{R}_u \left(x, y \right).
\end{align}

Contrasting this property with that in Equation~\ref{eq:lin}, it can be readily appreciated that uniform comparisons are invariant to feature translation, but not invariant to simultaneous scaling of the arguments.

A \emph{proportional comparison} should satisfy the following relationship:
\begin{align} \label{eq:proportional_comparison}
  \mathcal{R}_p \left(\alpha x, \alpha y \right) = 
     \mathcal{R}_p \left(x, y \right)
\end{align}

Unlike the uniform case, proportional comparisons are invariant to simultaneous scaling of the two arguments.

An example of proportional comparison between two generic variables $x$ and $y$, with $x\neq 0$, is:
\begin{align}\label{eq:exprop}
   \mathcal{R}_p \left(x,y \right) = \frac{y}{x} 
\end{align}

Indeed, we have that:
\begin{align}
   \mathcal{R}_p \left(\alpha \, x, \alpha \, y \right) = \frac{\alpha \, y}{ \alpha \, x} = \frac{y}{x} = \mathcal{R}_p \left(x, y \right)
\end{align}

Now, let us consider the relationship between the proportional comparisons $\mathcal{R}_p \left(y_{x_1},y_{x_2} \right)$ and $\mathcal{R}_p \left(y_{x_1+\alpha}, y_{x_2 + \alpha} \right)$ between pairs of proportional features obtained by transformation of free variables as $y_{x_1} = f(x_1)$, $y_{x_2} = f(x_2)$, $y_{x_1+\alpha} = f(x_1+\alpha)$, $y_{x_2+\alpha} = f(x_2 +\alpha)$. It follows from the property of proportional features expressed in Equation~\ref{eq:beta} that :
\begin{align}
   \mathcal{R}_p \left(y_{x_1+\alpha}, y_{x_2+\alpha} \right) = 
   \mathcal{R}_p \left(y_{\alpha} \, y_{x_1}, y_{\alpha} \, y_{x_2} \right) = \nonumber \\
   = \frac{y_{\alpha} \, y_{x_2}} {y_{\alpha} \, y_{x_1}} = 
   \mathcal{R}_p \left( y_{x_1}, y_{x_2} \right) \label{eq:relprop}
\end{align}

In summary, we have that:
\begin{align}
  \mathcal{R}_p \left(y_{x_1},y_{x_2} \right) = \mathcal{R}_p \left(y_{x_1+\alpha}, y_{x_2 + \alpha} \right)
\end{align}

which holds only if \emph{both} the features and comparisons are proportional.

The above developments and properties can be combined to yield:
\begin{align}
\boxed{
\begin{array}{ccc}
   \mathcal{R}_p \left(y_{x_1},y_{x_2} \right) & = & \mathcal{R}_p \left(\beta \, y_{x_1}, \beta \, y_{x_2}\right) \\
     \updownarrow & & \updownarrow \\
   \mathcal{R}_u(x_1,x_2) & = &\mathcal{R}_u(x_1+\alpha,x_2+\alpha)
\end{array} } \label{eq:rel}
\end{align}

where $\beta=c^{\alpha}$, $x_1$ and $x_2$ are uniform features, and $y_{x} = f(x)$ is a proportional transformation, and the arrows indicate transformations from uniform to proportional features and vice-versa.

\begin{figure}
  \centering
   \includegraphics[width=.9 \textwidth]{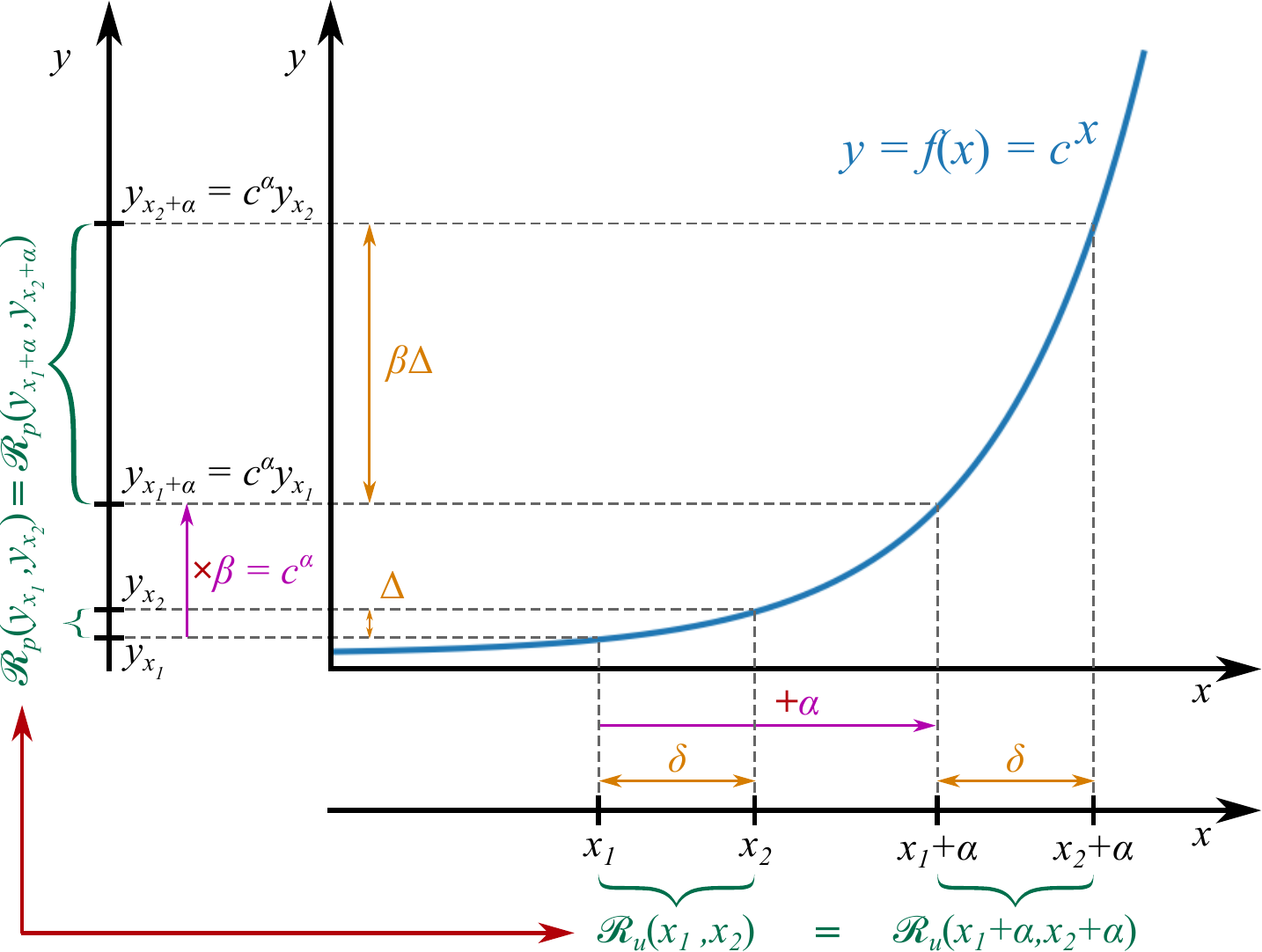}\\
 \caption{Graphical representation of the duality relationship (Eq.~\ref{eq:rel}) between the uniform and proportional spaces of features and comparisons, established by the transformation function $y=f(x)=c^x$. In case relations between comparisons performed in the uniform domain are to be preserved, proportional comparisons should be adopted in the transformed domain in order to compensate for the effect of the proportional transformation.}
 \label{fig:duality}
\end{figure}

Figure~\ref{fig:duality} illustrates the above duality relationship in graphical terms. Here, we have two pairs of uniform variables --- namely $x_1$; $x_2$ and $x_1+\alpha$; $x_2\alpha$ being transformed by a proportional function $f(x) = 2^x$.

The result in Equation~\ref{eq:rel} is particularly important because it indicates the consistency between uniform comparisons of translated uniform features and proportional comparisons of the counterpart scaled proportional features. This result also emphasizes the key aspects of translation invariance of comparisons in the uniform domain being intrinsically associated with scaling invariance of the comparisons in the proportional domain.

In section~\ref{sec:norm}, we shall consider feature normalizations particularly adequate to be applied in each of these two domains.

\subsection{Transforming and Relating Comparison Operations}

It should be observed that there is a virtually infinite number of possible uniform and proportional comparisons, a few of which have been discussed above. Each of these potentially infinite comparisons can possibly have distinct properties, such as regarding the sharpness of their comparison, and range of magnitude of the resulting comparison values.

The interesting aspect observed above can be better appreciated by the fact that, given a proportion (or even uniform) comparison $\mathcal{R}_p$, it immediately follows that:
\begin{align}
    \tilde{\mathcal{R}}_p = \left[ \mathcal{R}_p \right]^D
\end{align}
is also a proportional comparison, with $D>0$ being a real-valued constant. The incorporation of the exponent $D$ in the Jaccard similarity index discussed in the next section constitutes a particular example of the transformed comparisons approach.

It should be observed that the above \emph{comparison transformation} needs to satisfy some requirements, including the preservation of the \emph{order relation} established by the comparison operation to be transformed, an aspect which is related to the monotonicity of the transformation function.

Another interesting possibility consists of \emph{relating comparisons}. Let us consider a feature $x$, its version $y$ obtained by a function $y =f(x)$, as well as an arbitrary uniform comparison $\mathcal{R}_u$ and an arbitrary proportional comparison $\mathcal{R}_p$. Though the duality relationship expressed in Equation~\ref{eq:rel} holds, it does not immediately lead to a possible manner of relating the uniform and proportional comparison operations, such as:
\begin{align}
  \mathcal{R}_p(y_1,y_2) = g\left( \mathcal{R}_u(x_1,x_2) \right),
\end{align}
where $x_1$ and $x_2$ are arbitrary values of the variable $x$, and $g()$ is some generic function.

However, it is still possible to try identifying the function $g()$ by some other means. For instance, in the particular case of $f(x)=2^x$, the uniform comparison implemented in terms of the uniform comparison $\mathcal{R}_u(x_1,x_2) = x_2-x_1$ can be related to the proportional comparison indicated in Equation~\ref{eq:exprop} in the following manner:
\begin{align}
  \frac{y_2}{y_1} = 2^{\left( x_2-x_1 \right)} \longrightarrow
  \mathcal{R}_p(y_1,y_2) = 2^ { \mathcal{R}_u(x_1,x_2) }
\end{align}
so that, in this particular case, we would have $g(z) = 2^z$.

The above example helps understanding why the duality relationship does not immediately lead to a relationship between the uniform and proportional comparisons: it is also necessary to take into account the specific form of the comparison function (in this case difference and ratio), as well as the specific type of feature transformation $y=f(x)$ (in this case the choice of basis $2$ adopted in $y=2^x$). The range of magnitudes of the comparison values and respectively adopted normalizations should also be taken into account.

In case the function $g()$ can be obtained in the case of a specific situation, it could be used to complement the information provided by the duality relationship in Equation~\ref{eq:exprop}, as it establishes a direct relationship between implementing uniform comparisons in the uniform domain and calculating proportional comparisons in the respective proportional feature space.

Another prospect regarding comparison operations consists of combining them (e.g.~\cite{da2021further}), which is particularly interesting when the comparisons operations yield bound results, especially in the interval $[0,1]$. One of the ways in which two comparison operations can be combined is in terms of their product, as in:
\begin{align}
  \mathcal{R}_3(x_1,x_2) = \mathcal{R}_1(x_1,x_2) \, \mathcal{R}_2(x_1,x_2),
\end{align}
where a new comparison operation $\mathcal{R}_3(x_1,x_2)$ is obtained by the product of two previously available comparison operations. In case the preliminary operations are bound within the interval $[0,1]$, the new comparison will also be bound within that same interval. Observe also that a more strict comparison will be obtained. The coincidence similarity index to be described in the following section constitutes an example of this type of combination.

\subsection{Multiset-Based Proportional Comparisons}

Going back to Equation~\ref{eq:exprop}, we generally have that $\mathcal{R}_p(x,y) \neq \mathcal{R}_p(y,x)$. A related proportional comparison which is also \emph{commutative} can be obtained as:
\begin{align}
  \mathcal{R}_p \left(x,y \right) = \mathcal{R}_p \left(y,x \right) = \frac{\min(x,y)}{\max(x,y)}
\end{align}
with $0 \leq \mathcal{R}_p \left(x,y \right) \leq 1$, and $x,y \geq 0$.

Interestingly, this comparison operation can be verified to correspond to the multiset (e.g.~\cite{da2021multisets}) generalization of the Jaccard similarity index (e.g.~\cite{Jac:wiki}) for non-negative scalar values. It can be verified that this similarity index implements a proportional comparison between the scalars $x$ and $y$.

A multivariate version of the above similarity index can be obtained can be defined to compare two non-zero vectors $\vec{x}=[x_i]$ and $\vec{y}=[y_i]$ with non-negative entries and dimension $M$ as follows:
\begin{align}
  \mathcal{R}_p \left(\vec{x},\vec{y} \right) = \mathcal{R}_p \left(\vec{y},\vec{x} \right) = \frac{ \sum_{i=1}^{M} \min(x_i,y_i)}{ \sum_{i=1}^{M} \max(x_i,y_i)}
\end{align}
with $0 \leq \mathcal{R}_p \left(x,y \right) \leq 1$.

Observe that this multivariate comparison index is proportional, being invariant to a common scaling of all the variables, i.e.:
\begin{align}
  \mathcal{R}_p \left(r \vec{x},r \vec{y} \right) = \mathcal{R}_p \left(\vec{x},\vec{y} \right) 
\end{align}

where $r \in \mathbb{R}$, $r \neq 0$.

The Jaccard similarity index generalized (e.g.~\cite{da2021further,costa2022simil,costa2023mneurons}) to the comparison of two non-zero real-values vectors $\vec{x}$ and $\vec{y}$ with possibly negative entries, which are represented as np-sets~\cite{Costanpset}, can be expressed as:
\begin{align} \label{eq:Jacc}
   \mathcal{J}(\vec{x},\vec{y}) =& \, \mathcal{J}(\vec{y},\vec{x}) =
 \frac{|\vec{x} \cap \vec{y}|}{|\vec{x} \cup \vec{y}|} =
   \nonumber \\
   & = 
   \frac{ \sum_{k=1}^{M} \left[ \min \left( m^p_{x,k}, m^p_{y,k} \right) + \min \left( |m^n_{x,k}|, |m^n_{y,k}| \right) \right]}
   { \sum_{k=1}^M \left[ \max \left( m^p_{x,k}, u^p_{y,k} \right) + \max \left( |m^n_{x,k}|, |m^n_{y,k}| \right) \right] }.
\end{align}

where we have that the two original vectors to be compared are represented in terms of the respective np-sets~\cite{Costanpset}, i.e.:
\begin{align}
   &\vec{x} \longrightarrow \left\{ [1, m^p_{x,1}, m^n_{x,1}]; [2, m^p_{x,2}, m^n_{u,2}]; \ldots; [M, m^p_{x,M}, m^n_{x,M}] \right\} \\
   &\vec{y} \longrightarrow \left\{ [1, m^p_{y,1}, m^n_{y,1}]; [2, m^p_{y,2}, m^n_{v,2}]; \ldots; [M, m^p_{y,M}, m^n_{y,M}] \right\}
\end{align}

where:
\begin{align}
  &m^p_{x,k} = \max \left(x_k, 0 \right) \\
  &m^n_{x,k} = \min \left(x_k, 0 \right) \\
  &m^p_{y,k} = \max \left(y_k, 0 \right) \\
  &m^n_{y,k} = \min \left(y_k, 0 \right) 
\end{align}

It can be shown that $0 \leq \mathcal{J}(\vec{x},\vec{y}) \leq 1$.

It has been shown~\cite{da2021further} that the Jaccard index does not take into account the interiority index (also called overlap, e.g.~\cite{vijaymeena2016a}). As such, pairs of vectors with distinct interiority can be mapped into the same value of the Jaccard similarity index. Thus, it is interesting to consider the following np-set version of the \emph{interiority index}, capable of quantifying the relative interiority of the two compared vectors:
\begin{align}
   &\mathcal{I}(\vec{x},\vec{y}) =\mathcal{I}(\vec{y},\vec{x}) = \nonumber \\
   &= \frac{\sum_{k=1}^{M} \left[ \min \left( m^p_{x,k}, m^p_{y,k} \right) + \min \left( |m^n_{x,k}|, |m^n_{y,k}| \right) \right] }
   {\min \left( \sum_{k=1}^{M} \left[ m^p_{x,k} +|m^n_{x,k}| \right], \sum_{k=1}^{M} \left[ m^p_{y,k} +|m^n_{y,k}| \right] \right) }
   \label{eq:npcoinc}
\end{align}
with $0 \leq \mathcal{J}(\vec{x},\vec{y}) \leq 1$.

Observe that both the Jaccard and interiority similarity indices correspond to proportional comparisons.

The \emph{coincidence similarity index} generalized to real-valued, non-zero vectors $\vec{x}$ and $\vec{y}$, consists of the following combination of the Jaccard and Interiority indices:
\begin{align}
   \mathcal{C} (\vec{x},\vec{y}) = \left[\mathcal{J} (\vec{y},\vec{x}) \right]^D \ \left[ \mathcal{I} (\vec{x},\vec{y}) \right]^E \label{eq:coinc}
\end{align}
where $D$ and $E$ are exponents controlling the relative weight of the constraints implied by the interiority index. Again, we have that $0 \leq \mathcal{C} (\vec{x},\vec{y}) \leq 1$.

In Equation~\ref{eq:Jacc}, the sums are performed separately over minimum and maximum terms between features. It is also possible to define a modified version of the generalized Jaccard similarity index where the comparisons respective to each of the $M$ features are performed jointly (as, for instance, in the Canberra distance~\cite{lance1966}), i.e.:
\begin{align}
   \mathcal{J}_m(\vec{x},\vec{y}) =& \, \mathcal{J}_m(\vec{y},\vec{x}) =
   \nonumber \\
   & = \frac{1}{M} \sum_{k=1}^{M}
   \frac{ \min \left( m^p_{x,k}, m^p_{y,k} \right) + \min \left( |m^n_{x,k}|, |m^n_{y,k}| \right) }
   { \max \left( m^p_{x,k}, u^p_{y,k} \right) + \max \left( |m^n_{x,k}|, |m^n_{y,k}| \right) }
   \label{eq:Jaccmod}
\end{align}

where the normalizing term $\frac{1}{M}$ has been incorporated in order to imply that $0 \leq \mathcal{\tilde{J}}(\vec{x},\vec{y}) \leq 1$, since each of the $M$ separated comparisons can have one as maximum value. In addition, terms in the above sum that are equal to $\frac{0}{0}$ can be substituted by the value 1 (that would follow from understanding that the similarity between $0$ and $0$ is equal to $1$.).

As the interiority index has no effect in the case of comparisons between scalar values, it has not been considered. As we shall see in Section~\ref{sec:norm}, the modified Jaccard index has some particularly interesting properties regarding intrinsic normalization of proportional features.

\begin{figure}
  \centering
    \includegraphics[width=1 \textwidth]{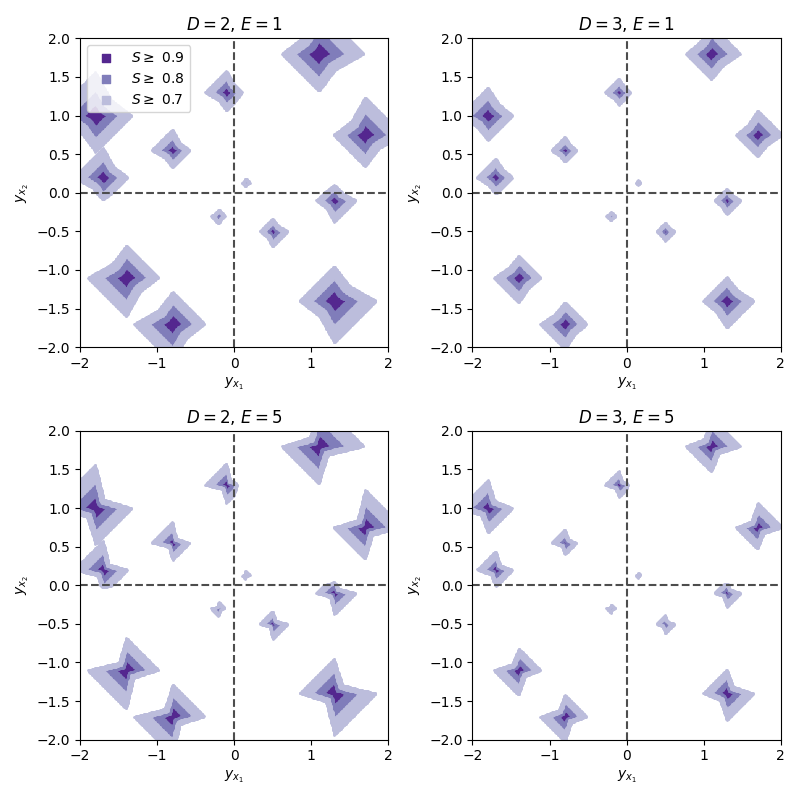}\\
 \caption{Graphic illustration of two-dimensional receptive fields obtained for the \emph{multiset coincidence similarity index} considering combinations of $D=2, 3$ and $E=1, 5$.}
 \label{fig:ex_recep}
\end{figure}

Figure~\ref{fig:ex_recep} illustrates two-dimensional receptive fields (corresponding to the regions leading to similarity values larger or equal to 0.7) of the multiset coincidence similarity index in Equation~\ref{eq:coinc} obtained for combinations of $D=2,3$ and $E=1,5$.

As expected, the size of the receptive fields decreases steadily with their distance from the center of the coordinate system. While the parameter $D$ controls the size of the receptive field, the parameter $E$ defines the respective lateral indentation.

\begin{figure}
  \centering
    \includegraphics[width=1 \textwidth]{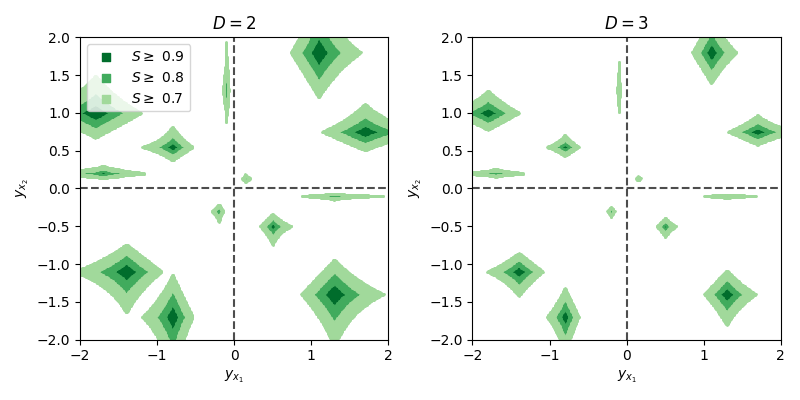}\\
 \caption{Graphic illustration of two-dimensional receptive fields obtained for the \emph{modified multiset Jaccard similarity index} considering combinations of $D=2,3$.}
 \label{fig:ex_recepmod}
\end{figure}

Examples of receptive fields characterizing the modified Jaccard similarity index for $D=2,3$ are depicted in Figure~\ref{fig:ex_recepmod}.

\subsection{A Note on Transformed Features}

In practice, it is not uncommon to have features that correspond to transformations/normalizations of respective previous features. Actually, it can be difficult to characterize an \emph{original reference} feature, since the model of the system from which the data originates is often not precisely known and also because measurements are often transformed in several manners in order to normalize values, remove noise, etc.

For instance, an originally proportional feature $x$ may have already been standardized into a new feature $\tilde{x}$ which, therefore, is no longer proportional. Ideally, it would be interesting to know as much as possible about the original features from which the supplied set of features have been obtained by respective transformations/normalizations because this knowledge could then allow the selection of particularly adequate respective normalizations.

While there are always choices between several units when measuring a given physical property, the type of and characteristics of this feature (e.g.~uniform, proportional, etc.) typically follow from the respective definition and modeling of the property itself, and not from the choice of units. For instance, the representation of the geometry (coordinates) of 2D or 3D objects almost invariably employs the Euclidean metric, therefore being uniform features irrespectively of the chosen units. As such, these coordinates can be translated and rotated in an Euclidean feature space without changing the respective relative distances between coordinates. However, distance measurements between object points represented as coordinates will change with scaling.

Left skewed feature densities are more typically related to the eventual choice of logarithmic scales. Intrinsically, this type of feature would imply that, given a cluster of data elements, the variation (observation error or group dispersion) of the feature values would tend to decrease with the respective average. Left skewed features therefore demand specific approaches to their normalization and comparison, which is not addressed in the present work. 

In principle, samples from a proportional feature obtained from the transformation of a uniform density by a top concavity monotonic function $f(x)$ will have right skewed density involving only positive sample values. In addition, they will yield a straight line when displayed as a semi-log plot. More specifically, we have from Equation~\ref{eq:propintvl} that the density of a proportional (right skewed) feature obtained from a uniform feature defined along the interval $[a,b]$ through the function transformation $y=f(x)=c^x$ (implying $x = log(y) \, /log(c)$), with $c>1$, will necessarily be:
\begin{align}
   p_y(y) = \frac{1}{f'(x)} \, \frac{1}{b-a} = \frac{1}{c^x} \, \frac{1}{\log(c) \ (b-a)} = 
 \frac{1}{c^{\frac{\log(y)}{\log(c)}}} \, k =k \, y^{-1},\label{eq:exp}
\end{align}
where $x \in [a,b]$, $y \in [c^a, c^b]$ and $k=1/\log(c)/(b-a) > 0$.

Therefore, by taking the logarithm of the above expression, it follows that:
\begin{align}
   \log(p_y(y)) = \log(k) \, - \log(y)
\end{align}

The above expression means that, when represented as a semi-log plot, it will yield a straight line with a slope equal to $-1$, indicating a scale-free density originating from the power law relationship. This suggests a possible approach that could, in principle, be considered for determining whether a given feature is proportional. However, this is not a straightforward endeavor.

In practice, the identification of the type of a given feature can be rather challenging or even impossible, which is often the case when the physical model of the features is not available. This procedure is complicated by the fact that the available samples could be too sparse, noisy, or cover only a small portion of the possible range of the feature. It may also happen that, though a given feature is obtained by a well-defined proportional transformation of another uniform feature, the available samples had been obtained from a skewed original density in the uniform domain.

\begin{figure}
  \centering
    \includegraphics[width=0.75 \textwidth]{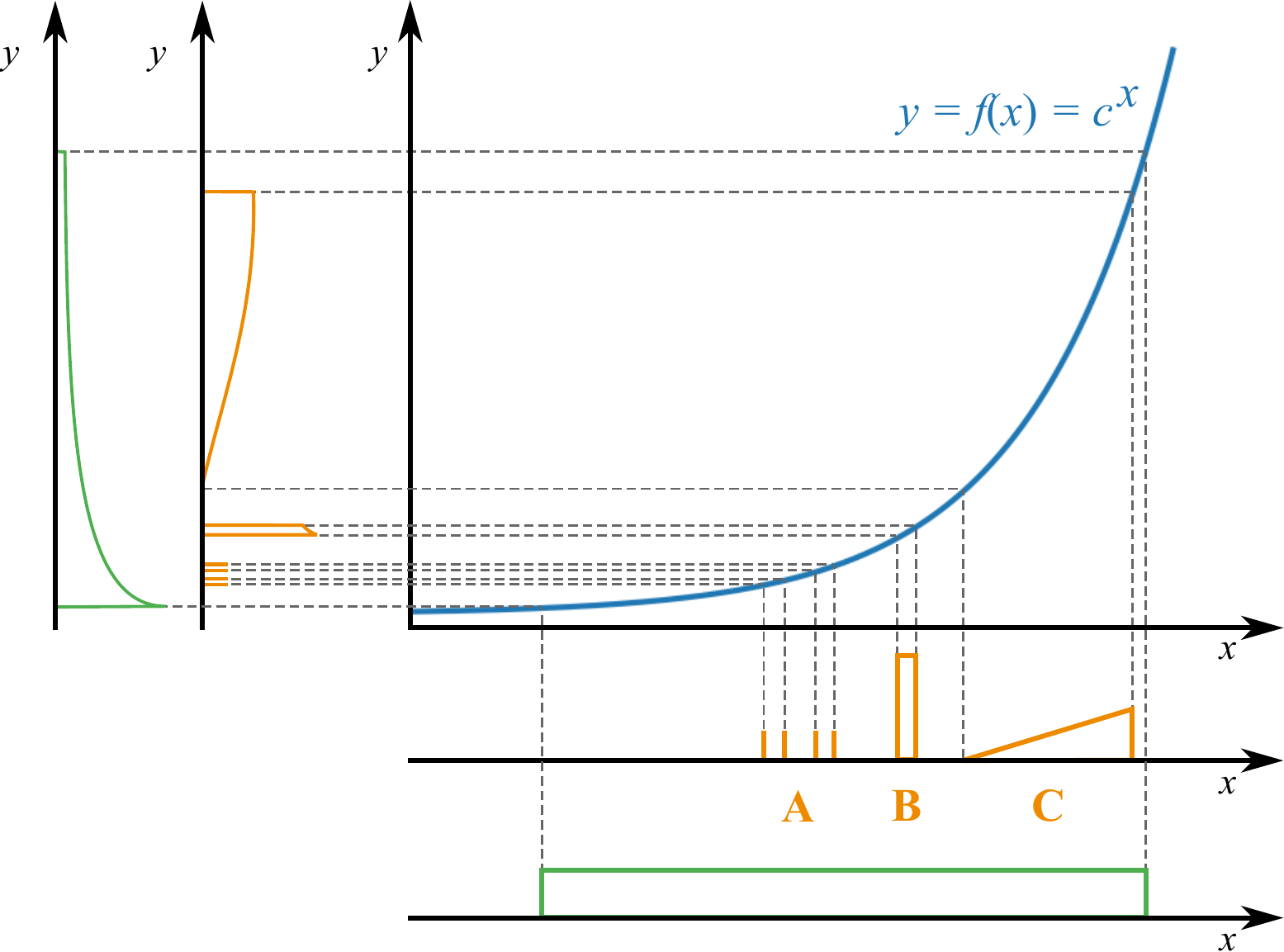}\\
 \caption{Graphic illustration of a situation in which, though a feature $y$ is obtained by a well-defined proportional transformation (namely $y= c^x$) of a uniform features $x$, the available samples are not enough for proper identification of the type of $y$. Also included (in green), for the sake of comparison, are a long uniform density and its respective transformation (see Eq.~\ref{eq:exp}). The densities magnitudes are now shown in absolute terms.}\label{fig:ex_challenges}
\end{figure}

Some of the above difficulties are illustrated in Figure~\ref{fig:ex_challenges}: though $y$ is a proportional feature derived from another variable $x$ by the proportional transformation $y=c^x$, the three available sample groups are limited as follows: (A) is too sparse; (B) only covers a narrow range of possible feature values; and (C) is biased (left skewed). The inference of $f(x)$ from any of these three groups taken separately or combined will very likely turn out to be incorrect.

Though it would be interesting to approach more formal methods that could be applied to try to infer the type of given features, this interesting possibility is not further investigated in the present work. 

Results obtained from data analysis and modeling should always be treated as being preliminary and subjected to further validation and enhancements.

\section{Normalization Methodologies} \label{sec:norm}

As discussed in Section~\ref{sec:integrated}, the choice of a suitable \emph{normalization procedure} to be applied into a dataset derived from some entities represented in terms of a given feature vector depends strongly not only on the characteristics of the original dataset and respectively adopted features but also on each type of intended analysis and modeling of the studied entities, as well as on the eventual normalization of the original features.  Specific aspects may also be implied by the questions and requirements underlying each specific problem being addressed.

\emph{Feature normalization} constitutes a particularly important and challenging issue in data characterization, comparison, analysis, classification, and modeling. Normalization is often required to avoid some features with intrinsically larger magnitudes predominating in the respective data analysis. Thus, given a set of features with varying magnitude extensions, some respective normalization procedures can be applied so as to obtain new features with similar, or more balanced/equalized magnitude variations. As such, this procedure can be understood as a particular case of \emph{feature transformation}. In principle, but not necessarily, the normalization of a feature could be expected not to change its form of statistical density, acting linearly on the magnitudes.

The key point addressed in the present work concerns the fact that the choice of a suitable normalization procedure needs to take into account the type and properties of the original features, as well as the methodologies to be subsequently applied for respective comparison, analysis, classification, and modeling. In principle, it could be expected that the normalization will not change the type of data, meaning that in case a feature is uniform or proportional, it will continue to be so after normalization. However, this is not mandatory, as in cases in which the normalization is supposed to equalize not only the magnitudes, but also act non-uniformly along the domain (e.g.~in order to zoom on specific ranges of feature magnitude).

In the present work, we shall focus our attention on the two following types of original features: (i) \emph{uniform features} to be compared in terms of \emph{uniform} comparisons; and (ii) \emph{proportional features} to be compared in terms of \emph{proportional} comparisons. For generality's sake, situations involving skewed features other than the proportional type are also considered. These three important cases are addressed in the following subsections, respectively.

In principle, in case all the available features have the same type, it is inherently more interesting to keep them in the same original domain, as this will facilitate the analysis and interpretation of related results. Datasets may also eventually involve heterogeneous features, for instance, combining uniform and proportional features. In these cases, a possible approach would be to transform between feature types, such as transforming uniform features in order to obtain all proportional features.

Yet another important point to be taken into account while implementing feature normalizations concern whether each of the involved features is treated separated as independently or jointly. Here, we shall be limited to the more common normalization approach (e.g.~as adopted in standardization) in which each feature is normalized independently of the others, as if we were dealing with a univariate density.

\subsection{Uniform Features}

A first important step while addressing the normalization of uniform features to be uniformly compared consists in taking into account the \emph{properties} of both uniform features and comparisons.

We have from the own definition of uniform features adopted in the present work that a uniform comparison between values of this type of feature should be invariant to \emph{feature translation}. This important intrinsic property suggests that a first step in respective normalization could consist of translating all feature values to the respective average (also corresponding to the center of mass). This can be formally written as:
\begin{align}
   \hat{x}_{i,k} = x_{i,k} - \mu_{x_k}.
\end{align}

The immediate effect of this feature transformation is that $\mu_{\hat{x}_k}=0$.

We already know that this transformation will have no impact on the uniform comparisons between the same pairs of original and transformed feature values. Now, in order to make the magnitude variation of each of the involved feature to become similar, it is possible to divide the values of the transformed uniform feature $\hat{x}_k$ by the respective dispersion quantified in terms of the associated standard deviation:
\begin{align}\label{eq:stand}
   \tilde{x}_{i,k} = \frac{\hat{x}_{i,k}} {\sigma_{\hat{x}_k}} = 
   \frac{x_{i,k} - \mu_{x_k}}{\sigma_{x_k}}
\end{align}
where it has been taken into account that the standard deviation of a set of values does not change when these values are translated by a same amount. It will follow that:
\begin{align}
  &\mu_{\tilde{x}_k} = 0; \\ 
  &\sigma_{\tilde{x}_k}=1.
\end{align}
This result confirms that the dispersion of all features will become equal, so that none of them will tend to predominate when comparing of the entities. The feature transformation indicated in Equation~\ref{eq:stand} coincides with the \emph{standardization operation} frequently adopted for the purpose of features normalization.

\subsection{Proportional Features}\label{sec:propnorm}

One first important fact to be considered when normalizing proportional features prior to proportional comparisons is that \emph{neither of these concepts are invariant to translation}.
Because translations can modify the proportionality of features, it is particularly interesting o implement the normalization of a proportional feature while maintaining the origin of the features coordinate axes. However, their magnitudes can be nevertheless equalized. One important intrinsic property of a proportional comparison between two scalars is that it is invariant to respective scaling, i.e.:
\begin{align}\label{eq:prop}
   \mathcal{R}_p \left(r \, y_1, r \ y_2 \right) = \mathcal{R}_p \left(y_1, y_2 \right),
\end{align}
where $r$ is a non-zero constant real value.

Therefore, scaling becomes intrinsically adequate for separately normalizing the magnitudes of proportional features to be related by proportional comparisons.

Observe that, though proportional comparisons between samples of a single feature are invariant to respective joint scaling by a single constant, comparisons between the set of values organized into feature vectors will be characterized by each feature being scaled by a possibly distinct value. That is the case when features with distinct magnitude variations are scaled by respective possibly distinct normalization constants, as adopted in standardization.

Henceforth, the total number of positive and negative samples is henceforth indicated as $N_p$ and $N_n$, respectively. The following measurement of absolute dispersion of the values of proportional feature $y_k$ from a feature vector $\vec{y}$, with $k=1,2,\ldots, M$, involves only scaling and no translation:
\begin{align}
   &\xi_{y_k,p} = \sqrt{ \frac{1}{N_{y_k,p}} \sum_{i=1}^{N_{y_k,p}} y_{i,k,p}^2 }, \\
   &\xi_{y_k,n} = \sqrt{ \frac{1}{N_{y_k,n}} \sum_{i=1}^{N_{y_k,n}} y_{i,k,n}^2 }.
\end{align}

When the features densities are to be considered instead of the respective samples, the above absolute expressions can be rewritten in terms of the following adaptations of the second non-central statistical moments:
\begin{align}
   &\xi_{y_k,p} = \sqrt{ \int_{0}^{\infty} y_K^2 \, p(y_k) \, dy_k }, \\
   &\xi_{y_k,n} = \sqrt{ \int_{-\infty}^{0} y_K^2 \, p(y_k) \, dy_k }
\end{align}

Because the variation of the negative and positive values of each proportional feature are not necessarily symmetric, it is sometimes interesting to normalize these two sets of values separately. In this case, the normalization of a proportional feature $y$ can be implemented only in terms of scaling as in the following:
\begin{align}
   &\tilde{y}_{i,k,p} = \frac{y_{i,k,p}} {\xi_{x,p}},\\ 
   &\tilde{y}_{i,k,n} = \frac{y_{i,k,n}} {\xi_{x,n}}. 
\end{align}

It can be shown that it will necessarily follow that:
\begin{align}
   &\xi_{\tilde{y}_{k,p}} = 1,\\ 
   &\xi_{\tilde{y}_{k,n}} = 1. 
\end{align}

In addition, observe that the above normalization yields two new variables which are necessarily dimensionless. This normalization is henceforth called \emph{separated proportional normalization}, abbreviated as SPN.

The above type of normalization gives the same importance to both negative and positive values of the original proportional feature, and is intrinsically related to situations in which these two subsets are distinctly treated by the transformation $f(x)$. In case both positive and negative values of the feature are to normalized jointly by the same scaling, it is possible to modify the normalization procedure described above as:
\begin{align}
   &\tilde{y}_{i,k,p} = \frac{y_{i,k,p}} { \max(\xi_{y,n},\xi_{y,p})} \\ 
   &\tilde{y}_{i,k,n} = \frac{y_{i,k,n}} { \max(\xi_{y,n},\xi_{y,p})}
\end{align}

This normalization is henceforth called \emph{joint proportional normalization}, abbreviated as JPN.

Again, the resulting normalized features are dimensionless. However, only the largest absolute dispersion between negative or positive feature values is considered for the normalization, so that both these two sets of values are scaled by the same amount. It can be verified that, in this case:
\begin{align}
   &\xi_{\max(\tilde{y}_{k,n},\tilde{y}_{k,p})} = 1, \\ 
   &\xi_{\min(\tilde{y}_{k,n},\tilde{y}_{k,p})} \leq 1 
\end{align}

It should be kept in mind that the, though involving only scalings, above feature normalizations imply possibly distinct values of the Jaccard, interiority, and coincidence indices. However, interestingly the modified versions of the Jaccard index (Eqs.~\ref{eq:Jaccmod}) will not be modified by either the SPN or JPN approaches. Thus, those indices present the particularly interesting property of being intrinsically n normalized respectively when applied to proportional features.

\subsection{Other Types of Right Skewed Features}

Thus far in this work, attention has been focused on uniform and proportional features and comparisons, as well as their intrinsic duality (e.g.~Eq.~\ref{eq:rel}). In particular, we have seen that proportional features can be obtained by proportional transformations of uniform features, i.e.~$y = c^x$. However, in practice, there can be several features that, though sharing with proportional features the property of being right skewed, correspond to features obtained by other types of transformations such as:
\begin{align}\label{eq:pol}
  y = f(x) = x^q,
\end{align}
where q is some positive integer value. When applied on uniform free variables $x$, this transformation will generate a scale free density (power law).

Because this type of non-linear feature transformation is not proportional, it becomes interesting to verify under which conditions (if any) they behave under proportional comparisons. 

Let us consider the simple proportional comparison specified in Equation~\ref{eq:exprop}. In this case, we can write:
\begin{align}
  \mathcal{R}_p(y_{x_1},y_{x_2}) = \frac{y_{x_1}} {y_{x_2}} = \frac{x_1^q} {x_2^q}.
\end{align}

Provided $x_1,x_2 \gg \alpha$, we can consider the following approximation:
\begin{align}
   &\mathcal{R}_p(y_{x_1},y_{x_2}) = \frac{y_{x_1}} {y_{x_2}} = \frac{x_1^q} {x_2^q}
  = \left[ \frac{x_1} {x_2} \right]^q \approx \nonumber \\
  &\approx \left[ \frac{x_1+\alpha} {x_2+\alpha} \right]^q = 
   \frac{(x_1+\alpha)^q} {(x_2+\alpha)^q} =
   \mathcal{R}_p(y_{x_1+\alpha},y_{x_2+\alpha}),
\end{align}
which is an approximation of the property indicated in Equation~\ref{eq:prop}.

Thus, we have that when $x_1,x_2 \gg \alpha$ and $q$ are not too large, the proportional comparisons of right skewed features of the form $ y = f(x) = x^q$ behave in a similar manner to which they would in case they were proportional features.

A similar result can be verified for other proportional comparisons including the Jaccard and coincidence similarity indices. These important results allow us, in several cases but not generally, to implement comparison approaches between several right skewed features as if they were proportional. It should be nevertheless kept in mind that some types of features, such as those characterized by left skewed densities, cannot be approximated in terms of proportional approaches.

\section{Experiments and Discussion}

The henceforth reported approach to feature normalization studies involves considering two types of datasets, normalizing them, and them obtaining respective similarity networks.

In the case of proportional features, similarity networks are obtained considering the approach known as \emph{coincidence similarity networks}~\cite{da2022coincidence}. In this case, each data element is represented as a node, and the coincidence similarity values are used as weights between pairs of nodes. In the case of uniform features, similarity networks are obtained by calculating the Euclidean distances, identifying the maximum value, and then taking the complements.

The examples consider the following univariate and right skewed densities:
\begin{enumerate}
   \item $q_1(x_1)$: Sum of a normal density with average 5.0 and standard deviation 1.0 and a normal density with average -1.0 and standard deviation 0.2. 
   \item $q_2(x_1)$: Sum of a normal density with average 5.0 and standard deviation 1.0 and a normal density with average -2.5 and standard deviation 0.2.
   \item $q_3(x_2)$: A normal density with average 3.0 and standard deviation 1.0. 
   \item $p_1(y_1)$: obtained from the density $q_1(x_1)$ respectively to the transformation $y_{x_1} = g(x_1)$.
   \item $p_2(y_1)$: obtained from the density $q_2(x_1)$ respectively to the transformation $y_{x_2} = g(x_1)$. 
   \item $p_3(y_2)$: obtained from the density $q_3(x_2)$ respectively to the transformation $y_{x_2} = g(x_2)$
\end{enumerate}

where:
\begin{align}
    y_{x} = g(x) = 
    \left\{
      \begin{array}{l l}
      2^x & \emph{for } \geq 0 \nonumber \\
      -2^{|x|} & \emph{for } x < 0
      \end{array}
    \right.
\end{align}

The univariate densities $y_1$ and $y_2$ are then combined in order to obtain the following two-variate densities defining categories A and B:
\begin{align}
  y_A(y_1,y_2) = p_1(y_1) \, p_3(y_2) \\
  y_B(y_1,y_2) = p_2(y_1) \, p_3(y_2)
\end{align}

\begin{figure}
  \centering
    \includegraphics[width=1 \textwidth]{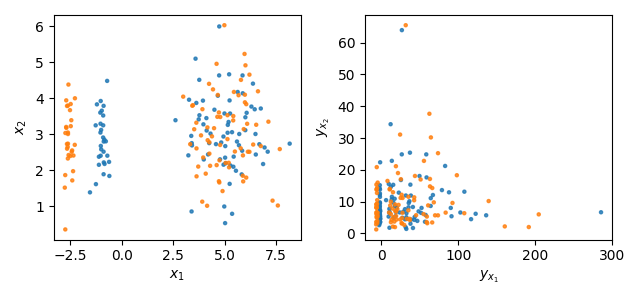}\\
    \hspace{.65cm} (a) \hspace{5.8cm} (b)
 \caption{Scatterplots illustrating the two-variate, two-categories dataset adopted in this section respectively to their uniform (a) and proportional (b) spaces. Categories A and B are shown in blue and orange, respectively.}
 \label{fig:scatt}
\end{figure}

Figure~\ref{fig:scatt} depicts the scatterplots respective sampled obtained from the uniform (a) and proportional (b) versions of the considered datasets.

It is interesting to observe that both categories are multimodal, in the sense of each presenting two separated groups (sum of gaussians). It should be kept in mind that, generally speaking, a category is not necessarily associated to a single respective cluster in its scatterplot. At the same time, the two categories in the adopted dataset are markedly similar, differing only as far as the two elongated clusters at the left-hand side of the scatterplots in Figure~\ref{fig:scatt}. Indeed, this characteristic has been specifically designed so as to constitute a challenge concerning the identification of the respectively involved modules and interconnection heterogeneities.

\begin{figure}
  \centering
    \includegraphics[width=1 \textwidth]{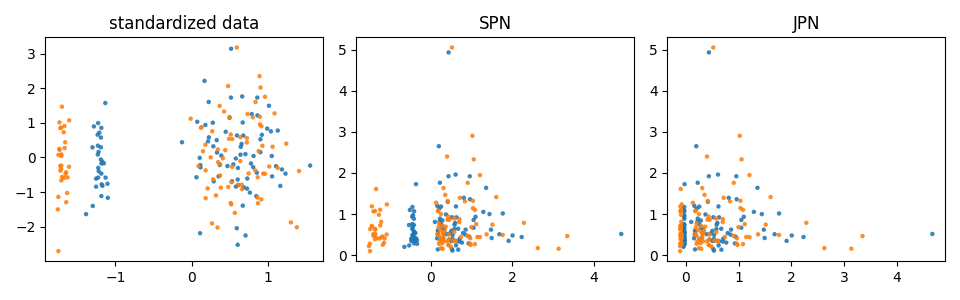}\\
    \hspace{.4cm} (a) \hspace{3.5cm} (b) \hspace{3.8cm} (c)
 \caption{Scatterplots illustrating the two-variate, two-categories dataset adopted in this section in normalized as follows: (a) standardization of uniform representation; (b) SPN of proportional representation; (c) JPN of proportional representation. Categories A and B are shown in blue and orange, respectively.}
 \label{fig:scattnorm}
\end{figure}

Figure~\ref{fig:scattnorm}(a) presents the standardized version of the considered dataset, while its SPN and JPN are shown in (b) and (c), respectively. It can be seen that the standardized data is centered at the origin of the coordinate axis, also presenting similar dispersions along the features $x_1$ and $x_2$. The proportional normalizations of the dataset have not been shifted but only scaled. Observe that the separated approach (SPN) yielded comparable dispersions along both the negative and positive values of the horizontal axis, while the JPN led to only the larger side (positive values) being normalized to unit dispersion.

\begin{figure}
  \centering
    \includegraphics[width=1 \textwidth]{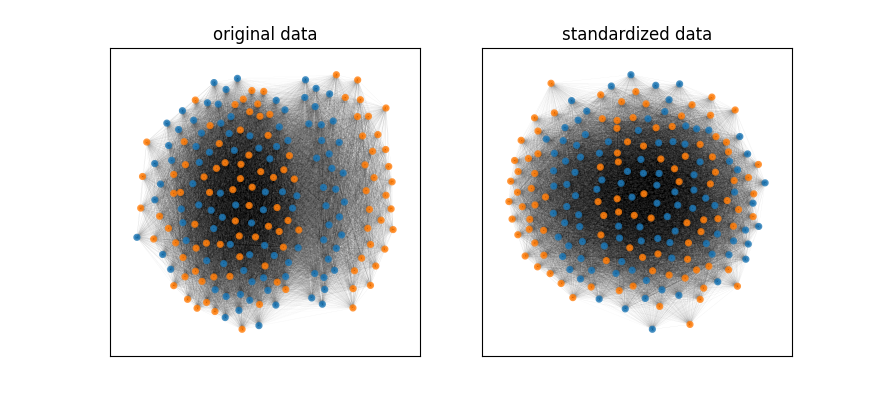}\\
    \hspace{.4cm} (a) \hspace{4.8cm} (b)
 \caption{Similarity networks obtained by using the complement of the Euclidean distance for the uniform version of the considered two-category dataset respectively to: (a) original data (non-normalized); and (b) standardized data. Similar results have been obtained.}
 \label{fig:net1}
\end{figure}

We start by considering the uniform version of the adopted dataset. Figure~\ref{fig:net1} shows the similarity network obtained for the uniform features by using the complement of the respective Euclidean distances while considering the non-normalized original data (a), as well as its standardized version (b). All similarity networks are henceforth visualized by using the Fruchterman-Reingold methodology (e.g.~\cite{fruchterman1991}).

Similar results have been obtained in which the several clusters in the original data are interconnected mostly in uniform manner. Though a band of blue points can be discerned, it is still strongly attached to the remainder of the network.

\begin{figure}
  \centering
     \includegraphics[width=1 \textwidth]{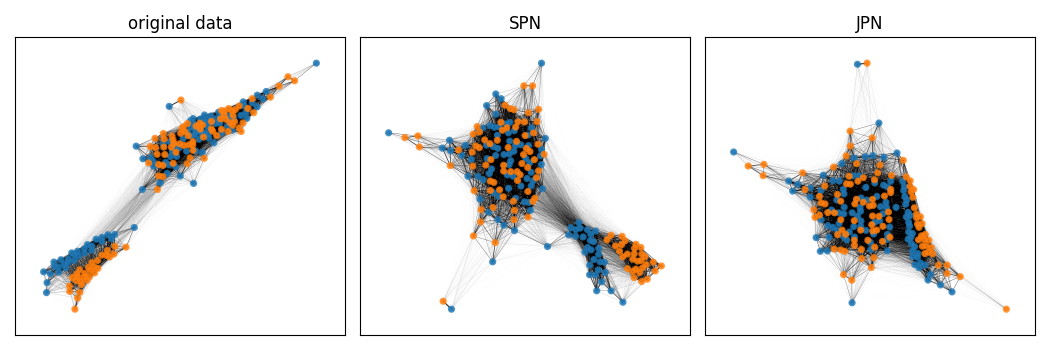} \\
     \hspace{.4cm} (a) \hspace{3.5cm} (b) \hspace{3.8cm} (c)
 \caption{Coincidence similarity (for $D=5$, $E=1$) networks obtained for the proportional version of the considered two-category dataset respectively to: (a) original data (non-normalized); (b) features normalized by using the SPN; (c) features normalized by using the JPN approach. The two colors identify the two categories. The SPN allowed the similarity network where the distinction between the two categories (orange and blue) is most emphasized.}
 \label{fig:net1_transformation}
\end{figure}

Figure~\ref{fig:net1_transformation} illustrates the coincidence similarity ($D=5$, $E=1$) between the proportional features of the considered dataset respectively to the non-normalized original data (a) as well as by the SPN (b) and JPN (c) approaches.

The coincidence networks obtained from the data features using the SPN approach emphasized the separation between the clusters in the original dataset. This interesting result is mainly a consequence of the most distinctive features of the two categories correspond to the negative feature values. Though these values presented smaller magnitudes than the positive samples, their importance was emphasized by the separated normalization approach underlying the SPN methodology. A difference between the results in Figure~\ref{fig:net1_transformation} concerns the fact that in (b) and (c), though the two more uniform modules are similar, the nodes in the larger heterogeneous module are more scattered than the network in (a). This interesting effect allows the interconnectivity within the larger, heterogeneous group, to be revealed in more details. 

\begin{figure}
  \centering
     \includegraphics[width=1 \textwidth]{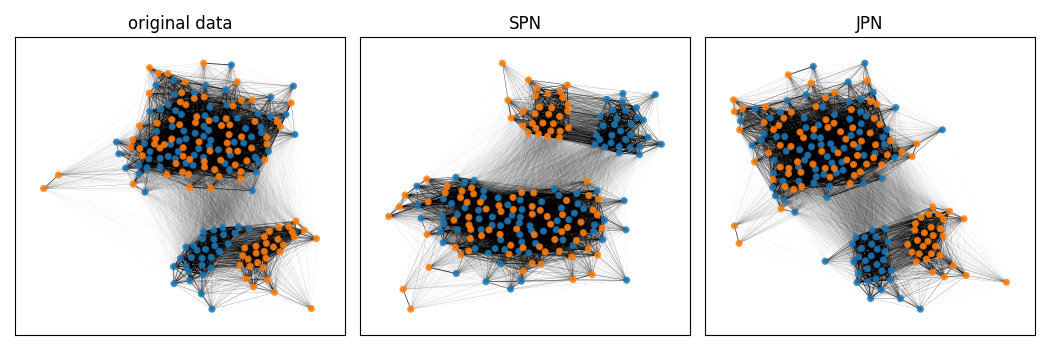} \\
     \hspace{.4cm} (a) \hspace{3.5cm} (b) \hspace{3.8cm} (c)
 \caption{Modified Jaccard similarity (for $D=5$) networks obtained for the proportional version of the considered two-category dataset respectively to: (a) original data (non-normalized); (b) features normalized by using the SPN; (c) features normalized by using the JPN approach. The two colors identify the two categories.}
 \label{fig:net1_modtransformation}
\end{figure}

Figure~\ref{fig:net1_modtransformation} presents the similarity networks obtained from the same dataset adopted above, but now using the modified Jaccard similarity index (Eq.~\ref{eq:Jaccmod}) instead of the multiset coincidence similarity index (Eq.~\ref{eq:coinc}).

As expected, the JPN had no effect on the result, because the modified Jaccard similarity index is by construction invariant to scaling by a same constant. The result obtained by using the SPN, however, is distinct because of the positive and negative feature values are normalized in separate, using possibly distinct scalings. Overall, the results in Figure~\ref{fig:net1_modtransformation} are similar, or even improved, than those shown in Figure~\ref{fig:net1_transformation}, in the sense of being characterized by more separated modules and detailed interconnections in the case of the heterogeneous module. This result, however, is specific to the considered separable two-dimensional density (the features are statistically independent).

All the above results are specific to the dataset and parameter configurations considered in the current work and are therefore not immediately extensible to other situations. Indeed, the choice of normalization and comparison approaches is highly dependent on several factors, including specific questions, requirements, type of data, etc., so that, in general, each of the normalizations and comparisons is not preferable in an absolute manner.

\section{Concluding Remarks}

The subject of feature normalization constitutes one of the main challenges in the areas of data representation, comparison, visualization, analysis, classification, and modeling. In principle, feature normalization need to be considered when the involved features present varying ranges of magnitudes, typically leading to dimensionless corresponding features.

The present work addressed feature normalization from the perspective of not only of the considered types of features --- which can be uniform, proportional, or more generic right skewed, but also while taking into account uniform and proportional comparisons which are typically employed in subsequent data analysis. The concept of feature transformation has been adopted as a central resource for better understanding possible relationships between distinct types of features and comparisons, as well as for addressing feature normalization.
Several concepts, methods, properties, and results have been described and discussed. First, a fully consistent duality relationship has been obtained (summarized in Eq.~\ref{eq:rel}) between proportional comparisons of proportional features and uniform comparisons of uniform features. Among other things, this relationship indicates that proportional features obtained from uniform features can be intrinsically approached by using proportional comparisons.

After briefly addressing normalization as part of a broader framework underlying data analysis and modeling, this work followed by presenting and discussing uniform and proportional features and comparisons, as well as how more general right skewed features can be approximately addressed in terms of proportional comparisons, the key issue of features normalization could be approached in a principled manner while taking into account the respective types of features and comparisons to be employed. This approach led to the identification of two intrinsically interesting normalizations to be applied in the uniform and proportional cases, which correspond to the methodology commonly known as standardization and normalization based on non-centralized dispersion of the involved feature values. Both these normalizations lead to dimensionless versions of the original features, each of which presenting unit respective dispersion.

Experiments involving synthetic data have also been presented and discussed concerning the estimation of similarity networks from uniform and proportional features by considering uniform and proportional comparisons. The obtained results confirmed that distinct normalizations can significantly influence the respective subsequent analysis. In the specific considered dataset, involving two-categories characterized in terms of two features, the SPN approach allowed the best identification of the clusters, with little separation being observed when the data was approached from its uniform domain. In addition, a modified version of the multiset Jaccard, interiority and coincidence similarity indices has been described which is intrinsically normalized but which is intrinsically adapted to statistically independent features.

It should be kept in mind that the results and discussion in this work are specific to the considered datasets, assumptions, and configurations so that they cannot be immediately generalized or extended to other problems and situations. Actually, the normalization of datasets deserves special attention as it is highly dependent on several aspects, including the type and properties of the data and features, methods to be employed for subsequent comparison and analysis, as well as the specific context and main questions underlying each problem and application. 

Among the several research possibilities motivated by the described concepts, methods, results, and discussions, we have the consideration of left skewed features, other types of transformations (e.g.~implemented by monotone top concave functions) and datasets, to consider features in the presence of noise and/or distortions, as well as other types of comparisons and normalizations, as well as considering experiments with dimensions larger than 2.

\section*{Acknowledgments}
A. Benatti is grateful to MCTI PPI-SOFTEX (TIC 13 DOU 01245.010\\222/2022-44), FAPESP (grant 2022/15304-4), and CNPq. Luciano da F. Costa thanks CNPq (grant no.~307085/2018-0) and FAPESP (grant 2022/15304-4).

\bibliography{ref}
\bibliographystyle{unsrt}

\end{document}